\documentclass[runningheads]{llncs}
\usepackage{graphicx}
\newcommand{\mc}[1]{\multicolumn{1}{c}{#1}}
\newcommand{\dvp}{DLv$3+$}
\newcommand{\upr}{UPN}
\newcommand{\lov}{Lov\'{a}sz}
\newcommand{\tbf}[1]{\textbf{#1}}
\newcommand{\tick}{\checkmark}
\newcommand{\br}[1]{\textbf{\textcolor{red}{#1}}}

\newcommand{\bb}[1]{\textbf{\textcolor{blue}{#1}}}

\usepackage{hyperref}

\usepackage[super]{nth}
\usepackage{booktabs}
\usepackage{amssymb}
\usepackage{amsmath}
\usepackage{verbatim} 
\usepackage{authblk}
\usepackage{url}

\makeatletter
\renewcommand*{\@fnsymbol}[1]{\ensuremath{\ifcase#1\or *\or \dagger\or \ddagger\or
    \mathsection\or \mathparagraph\or \|\or **\or \dagger\dagger
    \or \ddagger\ddagger \else\@ctrerr\fi}}
\makeatother

\usepackage{tablefootnote}

\DeclareMathOperator{\EX}{\mathbb{E}}

\usepackage[colorinlistoftodos]{todonotes}          
\usepackage[binary-units]{siunitx}                  
\usepackage{graphicx}                               
\DeclareSIUnit\px{px}                               
\newcommand{\ta}[1]{\textbf{$t_{#1}$}}

\begin{document}

\title{Effective semantic segmentation in Cataract Surgery: What matters most?}

\author{Theodoros Pissas\inst{1,2}\thanks{The first two authors contributed equally.} \and
Claudio S. Ravasio\inst{1,2}\textsuperscript{*} \and \authorcr
Lyndon Da Cruz\inst{3,4}\thanks{The last two authors contributed equally.} \and
Christos Bergeles\inst{2}\textsuperscript{$\dagger$}}

\authorrunning{Pissas*, Ravasio*, et al.}

\institute{Wellcome/EPSRC Centre for Interventional and Surgical Sciences, University College London (UCL) \and
School of Biomedical Engineering \& Imaging Sciences, King's College London (KCL) \and
Moorfields Eye Hospital, London \and Institute of Ophthalmology, University College London}

\maketitle              
\begin{abstract}

Our work proposes neural network design choices that set the state-of-the-art on a challenging public benchmark on cataract surgery, CaDIS. Our methodology achieves strong performance across three semantic segmentation tasks with increasingly granular surgical tool class sets by effectively handling class imbalance, an inherent challenge in any surgical video. We consider and evaluate two conceptually simple data oversampling methods as well as different loss functions. We show significant performance gains across network architectures and tasks especially on the rarest tool classes, thereby presenting an approach for achieving high performance when imbalanced granular datasets are considered. Our code and trained models are available at \url{https://github.com/RViMLab/MICCAI2021_Cataract_semantic_segmentation} and qualitative results on unseen surgical video can be found at \url{https://youtu.be/twVIPUj1WZM}.
\end{abstract}

\section{Introduction}\label{intro}
Cataract is the leading cause of blindness \cite{WHO_blindness} while also being predominantly preventable through surgery \cite{wang2016cataract}. Surgical data science envisions the deployment of data-driven systems to enhance clinical practice planning, delivery and quality assessment \cite{SDSc2020}. Intraoperative scene understanding constitutes a necessary building block towards utilising such systems in the context of computer assisted interventions, as it encapsulates recognition problems ranging from image-level classification of the procedure undertaken at a specific time step to the semantic labelling of every pixel in the scene.

Within the domain of cataract surgery, various works have focused on phase and tool presence detection \cite{CATARACTS_workflow,Padoy2012,MoritaPhase,DeepPhase,PhaseJAMA}, estimating only global video-level or frame-level information. Furthermore, some research has addressed the task of tool recognition and localisation via bounding box estimation \cite{CataractTools2}, segmentation \cite{CataractTools1}, or both \cite{CaDIS_tool_segmentation}, extracting information about tools against an all-encompassing background class. Recent advancements in computer vision \cite{Deeplabv3plus,OCR,UPerNet} demonstrate deep convolutional nets to be powerful semantic segmentation models for natural scenes if provided with large, pixel-wise annotated datasets. Our work therefore goes further and aims to obtain a semantic segmentation of both tools and all anatomical regions in the scene. Importantly, different from \cite{CataractTools2,CataractTools1}, we report results on a publicly available dataset, CaDIS \cite{CaDIS}, thus enabling reproducibility.

A common characteristic of cataract surgery video datasets is class imbalance caused by the small size of tools relative to anatomical regions, and the overall sparse appearance of certain tools in the duration of a surgical procedure. Coupled with this is the second major challenge of inter-class resemblance: using a more granular set of tool classes further reduces the perceived variation in tool appearance along with the number of samples per class, thus increasing imbalance. Consequently, and as demonstrated in \cite{CaDIS}, fine-grained tool recognition and segmentation is a particularly challenging task, with increases in class granularity leading to significant performance drops.

Our paper addresses these challenges for pixel-level semantic segmentation of cataract surgery videos. Our contributions are the following:
\begin{itemize}
    \item We demonstrate that oversampling in the form of repeat factor sampling, previously only proposed for long-tail instance segmentation \cite{gupta2019lvis}, or a custom adaptive sampling algorithm lead to significant gains in performance, especially for the rarest classes and in tasks where the class distribution is most imbalanced.
    \item We conduct detailed ablation studies involving different network architectures, data sampling strategies, loss functions, and training policies.
    \item From this, we determine a training policy consisting of a data oversampling method, a loss function and a training schedule that achieves the highest performance across benchmark sub-tasks and network architectures tested.
\end{itemize}
Ultimately, our top-performing models significantly outperform the results reported in \cite{CaDIS} on all sub-tasks of the CaDIS dataset, setting the state-of-the-art on this challenging benchmark.

\section{Materials and Methods}\label{methods}
We first present the evaluated design choices spanning network architectures, loss functions and two data oversampling strategies.

\subsection{Data}\label{methods:data}
\begin{figure}[b]
    \centering
    \includegraphics[width=1\linewidth]{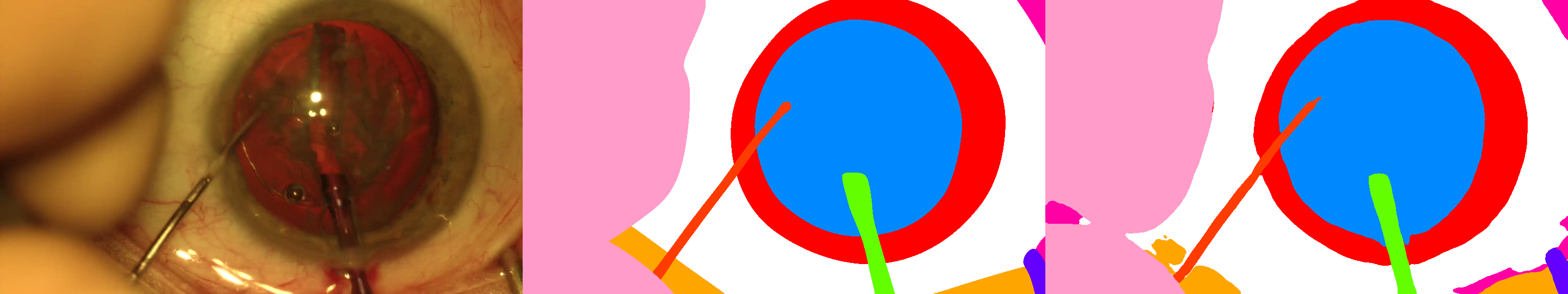}
    \caption{From left to right: An image from the CaDIS dataset (Video $22$, frame $5040$), its ground truth segmentation, and a sample prediction from our network for \ta3. The labels show the pupil (light blue), iris (red), cornea (white), skin (magenta), surgical tape (orange), an eye retractor (purple), and the surgeon's hand (pink). Two tools are present: a micromanipulator (orange red) and a phacoemulsifier handpiece handle (green).}
    \label{fig:data}
\end{figure}

Experiments were conducted using the public CaDIS dataset \cite{CaDIS} consisting of 25 surgical videos that contain segments of cataract surgery and comprise $4671$ pixel-level annotated frames with labels for \textit{anatomies}, \textit{instruments} and \textit{miscellaneous objects}. We follow \cite{CaDIS} that defined $3$, increasingly granular, semantic segmentation tasks. Task $1$ that entails $8$ classes: $4$ anatomies, $1$ all-encompassing instrument class and $3$ classes for miscellaneous objects. Task $2$ increases the number of classes to $17$ by splitting the all-encompassing instrument class of task $1$ to $9$ distinct classes allowing the separate recognition of different tools. Finally, task $3$ pushes instrument recognition granularity even further by considering the handles of certain instruments as separate classes, and has $25$ classes overall. In the remainder of the manuscript we refer to different tasks as \ta1, \ta2, and \ta3 for brevity. An example image can be seen in Fig.~\ref{fig:data}.

Data inspection \cite{karpathy_recipe} revealed $179$ at least partially mislabelled frames, most frequently affecting tool classes, which may obfuscate conclusions. We corrected the tool labels on $40$ of these frames, and excluded the rest from consideration. This reduced the overall dataset from $4670$ to $4491$ records. We use the same training set as Grammatikopoulou et al. \cite{CaDIS}, and report results on their original unfiltered train-val-test split to ensure reproducibility. However, the original test set \cite{CaDIS} is overly limited in size and no robust conclusions can be drawn: some of the rarest classes are present in just one of the three videos in the test set, with as few as seven highly correlated sequential instances in the $587$ frames overall. Therefore, we merge their validation and test sets for our more extensive evaluation. This ultimately yields $3550$ frames from $19$ subjects for training and $1120$ frames from $6$ different subjects for validation, or $3490$ training records and $1001$ validation records after correcting for mislabelled frames.

\subsection{Network Architectures}\label{methods:architectures}
We focus on networks with the classic encoder-decoder structure, using ResNets \cite{ResNet} as the standard encoder or network backbone. We made use of the implementations in the Torchvision library \cite{torchvision}, removing the final fully connected layers and passing the features to the decoder. As decoders or network heads we selected current state-of-the-art options which instantiate three different mechanisms for enhancing and decoding the encoder's feature maps into dense semantic labelling, namely a Feature Pyramid Network (FPN), an attention based contextual module and a dilated convolution context module.

The \tbf{UPerNet} (\upr{}) head \cite{UPerNet} consists of a FPN which takes the features from ResNet layers $2$--$5$ as input and combines them with a Pyramid Pooling Module (PPM) at the lowest resolution level. The outputs are fused into a single feature map, which is connected to a classifier that predicts the semantic segmentation at a quarter scale relative to the original input image. This prediction is then upsampled to the original resolution. We based our code on the implementation in \cite{zhou2017scene,zhou2018semantic}, with FPN and PPM channel lengths of $512$.

The \tbf{OCRNet} (OCR) head \cite{OCR} obtains a coarse intermediate prediction using a convolutional output head, operating on top of features of layer $4$ of the encoder. Together with layer $5$, this results in an aggregate feature representation for each object in the scene, fed to the Object Contextual Representation module to enhance it with contextual information based on object-pixel similarities. Subsequently, a convolutional layer maps them to the final prediction upsampled to the original resolution. We based our code on the official implementation \cite{OCR}, using $512$ channels for all convolutional layers. OCRNet uses a dilated ResNet backbone, replacing the stride $2$ convolutions in layers $4$ and $5$ (\cite{ResNet}) with a dilated convolution with rate $2$. The spatial dimensions of the encoder's output features result as $1/8$ of the input resolution, rather than the default $1/32$.

The \tbf{DeepLabv$3+$} (\dvp{}) head \cite{Deeplabv3plus} utilises an Atrous Spatial Pyramid Pooling (ASSP) module which encodes multi-scale context through $5$ parallel dilated convolutions with dilation rates $\{1, 6, 12, 18\}$. This is followed by a coarse-to-fine decoder which fuses layer $2$ of the encoder with the ASPP features to produce a prediction at a quarter of the input resolution which is then upsampled. We adapted the official Tensorflow implementation \cite{Deeplabv3plus} for Pytorch, using $256$ channels for all convolutions. \dvp uses the same dilated ResNet as OCRNet.

\subsection{Loss Functions}\label{methods:loss}
Cross-entropy as the standard first choice of loss function for semantic segmentation struggles to deal with strong class imbalances such as the ones encountered in \ta2 and \ta3, and minimising the cross-entropy often correlates poorly with maximising the actual performance metric, i.e. mean intersection over union (mIoU). Therefore, we explored a variation that uses online hard example mining (OHEM) \cite{ohem} by ignoring the loss in all pixels where the correct label is predicted with a probability above a threshold value, chosen as $0.7$. This follows the work on detection in \cite{ohem}, and we used an implementation from \cite{ohem_code}. While directly training on the mIoU metric is intractable due to its non-differentiability, surrogates such as the \lov{}-Softmax have been developed. It makes use of the \lov{} extension of the Jaccard index to create a differentiable loss function whose optimisation corresponds more closely to minimising the intersection over union metric \cite{Lovasz}. We used the PyTorch implementation provided by the authors.

\subsection{Addressing class imbalance}\label{methods:imbalance}
The number of class instances in the dataset vary by three orders of magnitude, as does the average number of pixels per class \cite{CaDIS}. This extreme imbalance affecting the labels for \ta2 and especially \ta3 is the main hurdle to achieving a high mIoU on this dataset. We evaluated two different approaches to overcome this, aimed at increasing the frequency with which the network is presented with the rarer and therefore harder to learn labels. Repeat factor sampling (RF) relies on repeating specific records within the dataset, based on pre-computed image-level scores, increasing the number of records per epoch. Adaptive sampling on the other hand moves away from the concept of an 'epoch' as iterating through the dataset once, and instead follows an online stochastic selection process for each batch, no longer guaranteeing each record will be seen in every epoch.

\noindent\tbf{Repeat Factor Sampling:}
We tailored repeat factor sampling, introduced in \cite{gupta2019lvis} for instance segmentation under long-tailed class distributions, to the task of semantic segmentation. First, for each class $c$ we compute a class-level repeat factor as $r_c = \max(1,\sqrt{t/f_c})$ where $f_c$ is the frequency of image-level occurrence of this class and $t$ is a frequency threshold hyper-parameter. Then, for each image $I$ in the dataset, an image-level repeat factor is computed as $r_I=\max_{c \in I} r_c$, which specifies the number of times $I$ should be repeated during an epoch of training and in general is not an integer. The value of $r_I$ is stochastically rounded before each epoch resulting in a randomly varying number of oversampled images per epoch, while over the course of training $\EX[r^{rounded}_I] \approx r_I $. We select $t=0.15$ to oversample images with classes such that $f_c<0.15$.

\noindent\tbf{Adaptive Sampling:}
Adaptive sampling is based on the principle of using the current per-class performance, calculated as a moving average of IoU values achieved during training with an exponential smoothing factor of $0.1$, initialised at $0.5$ for numerical stability. Therefore, our method is biased towards the selection of whole images with a high incidence of globally underperforming classes, as opposed to Berger et al \cite{berger2018adaptive} who compile training batches selecting patches with high local error values within the image data. The IoU values, which measure class performance, are converted into percentages expressing how many records in the batch should be selected while prioritising each class: $p = softmax(1 - {IoU}^2)$. Each of these records is chosen as the the one with the highest number of pixels labelled with the class of interest out of $10$ random samples from the dataset. Thus, the selection is automatically biased in favour of a higher incidence of underperforming, often rarer classes while not restricting it to any specific subset of the data.

\subsection{Training schedule}\label{methods:training}
As default settings, the networks were trained using the Adam optimiser \cite{Adam} 
for a maximum of $50$ epochs, with each epoch including repeated data if repeat factor or adaptive sampling was used. We chose a batch size of $8$, images at their original resolution of $540 \times 960$, and online data augmentation. The latter consists of random horizontal flips with probability $0.5$, Gaussian blurring with a kernel size randomly chosen between $3$ and $7$ occurring with a probability of $0.05$, and the Torchvision `colorjitter' augmentation with brightness, contrast and saturation adjustments in $[2/3, 3/2]$, and hue adjustment in $[-0.05, 0.05]$.

\noindent\tbf{Learning Rate:}
Two learning rate decay functions were tested: the exponential formula $lr = lr_0 * \alpha^i$ and the polynomial formula $lr = lr_0 * (1 - i / n)^{p}$ \cite{polynomial}, where $lr$ is the current learning rate, $lr_0$ the initial learning rate, $\alpha$ and $p$ hyperparameters to be chosen, $i$ the current epoch number, and $n$ the total number of epochs. We also used restarts at \SI{65}{\percent} of the previous $lr_0$, following the proposal in \cite{polynomial}. Ultimately, we chose the exponential decay with $lr_0 = 10^{-4}$ and $\alpha = 0.98$ as standard for all experiments in this work, as the other options explored yielded no significant benefits.

\noindent\tbf{Initialisation:}
Decoders were randomly initialised, while we used Imagenet-pretrained weights for backbones. We also experimented with using MoCov$2$ weights from self-supervised pre-training on Imagenet \cite{MoCov2} for a ResNet$50$.

\subsection{Implementation Details}\label{methods:ware}
Our implementation uses PyTorch 1.6. Experiments were run on an NVIDIA Quadro P6000 GPU with \SI{24}{\giga\byte} memory for \upr{} models (around \SI{20}{\giga\byte} of memory, \SI{20}{\hour} training time), and an NVIDIA Quadro RTX 8000 GPU with \SI{48}{\giga\byte} memory for OCRNet (around \SI{23}{\giga\byte} of memory, \SI{17}{\hour} training time) and \dvp{} models (around \SI{24}{\giga\byte} of memory, \SI{25}{\hour} training time).

\section{Results and Discussion}\label{results}
The main results are collated in Table~\ref{tab:results}. The metric reported is the best mean Intersection over Union (mIoU) value achieved on the validation dataset over the course of the $50$ training epochs. As shown in Table \ref{tab:results}, combining any of the architectures with the \lov{} loss and RF (Sec.~\ref{methods:imbalance}) outperforms all other variants consistently across all three tasks \ta1, \ta2, \ta3, providing an effective training policy for effective semantic segmentation across different levels of class granularity. We also observed that a mean ensemble of the best models per head further improves performance on all tasks, as reported in Table~\ref{tab:results}. Following are our key conclusions for what matters for high performance in this benchmark.

\begin{table}[ht]
\setlength{\tabcolsep}{3pt}
  \centering
   \caption{Results for \ta1, \ta2, and \ta3. \tbf{Bold} numbers denote the best results per task and model head, $\dagger$ indicates MoCov2 initialisation, Ensemble uses models in bold per task.}
  \label{tab:results}
  \begin{tabular}{c c c c c c c c c c}
    \toprule
    \multicolumn{2}{c}{Model}        & \multicolumn{3}{c}{Loss}              & \multicolumn{2}{c}{Sampling} & \multicolumn{3}{c}{Mean IoU}  \\
    \cmidrule(r){1-2} \cmidrule(l){3-5}  \cmidrule(l){6-7} \cmidrule(l){8-10}
    ResNet     & Head                & CE           & OHEM        & \lov{}     & RF          & Adapt.     & \ta1         & \ta2         & \ta3 \\
    \mc{$50$}  & \mc{OCR}            & \mc{\tick}   & \mc{}       & \mc{}      & \mc{\tick}  & \mc{}      & \mc{0.8950} & \mc{0.8102} & \mc{0.7624} \\ 
    \mc{$50$}  & \mc{OCR}            & \mc{}        & \mc{\tick}  & \mc{}      & \mc{\tick}  & \mc{}      & \mc{0.8967} & \mc{0.8102} & \mc{0.7742} \\
    \mc{$50$}  & \mc{OCR}            & \mc{}        & \mc{}       & \mc{\tick} & \mc{}       & \mc{}      & \mc{0.8979} & \mc{0.8109} & \mc{0.7345} \\
    \mc{$50$}  & \mc{OCR}            & \mc{}        & \mc{}       & \mc{\tick} & \mc{\tick}  & \mc{}      & \mc{0.8999} & \mc{0.8236} & \mc{\tbf{0.7777}} \\
    \mc{$50^{\dagger}$} & \mc{OCR}   & \mc{}        & \mc{}       & \mc{\tick} & \mc{\tick}  & \mc{}      & \mc{\tbf{0.9013}} & \mc{\tbf{0.8282}} & \mc{0.7512} \\
    \mc{$50$}      & \mc{OCR}        & \mc{}        & \mc{}       & \mc{\tick} & \mc{}       & \mc{\tick} & \mc{0.8997} & \mc{0.8220} & \mc{0.7632} \\
    \midrule
    \mc{$34$} & \upr                 & \mc{\tick}   & \mc{}       & \mc{}      & \mc{\tick}  & \mc{}      & \mc{0.8957} & \mc{0.8013} & \mc{0.7534} \\
    \mc{$34$} & \upr                 & \mc{}        & \mc{\tick}  & \mc{}      & \mc{\tick}  & \mc{}      & \mc{0.8967} & \mc{0.8039} & \mc{0.7491} \\
    \mc{$34$} & \upr                 & \mc{}        & \mc{}       & \mc{\tick}      & \mc{}  & \mc{}      & \mc{0.8990} & \mc{0.8151} & \mc{0.7374} \\
    \mc{$34$} & \upr                 & \mc{}        & \mc{}       & \mc{\tick} & \mc{\tick}  & \mc{}      & \mc{\tbf{0.8992}} & \mc{\tbf{0.8298}} & \mc{\tbf{0.7735}} \\
    \mc{$34$} & \upr                 & \mc{}        & \mc{}       & \mc{\tick} & \mc{}       & \mc{\tick} & \mc{0.8988} & \mc{0.8198} & \mc{0.7457} \\
    \midrule
    \mc{$50$} & \dvp                 & \mc{}        & \mc{}       & \mc{\tick} & \mc{\tick}  & \mc{}      & \mc{\tbf{0.8996}} & \mc{\tbf{0.8250}} & \mc{\tbf{0.7763}} \\
    \mc{$50^{ \dagger}$} & \dvp      & \mc{}        & \mc{}       & \mc{\tick} & \mc{\tick}  & \mc{}      & \mc{0.8983} & \mc{0.8224} & \mc{0.7578} \\
    \midrule
    \multicolumn{2}{c}{Ensemble}     & \mc{}        & \mc{}       & \mc{\tick} & \mc{\tick}  & \mc{}      & \mc{0.9020} & \mc{0.8360} & \mc{0.7870} \\

    \bottomrule
  \end{tabular}
\end{table}

\noindent\tbf{The superiority of the \lov{}-Softmax loss:}
A consistent finding was that the \lov{}-Softmax loss outperforms cross-entropy based losses as the default option in the literature \cite{Deeplabv3plus,OCR,UPerNet}. Notably, adding the \lov{} in \ta1 and \ta2 already leads to performance that is on par with models using a combination of cross-entropy and oversampling, the latter increasing training time significantly in the case of RF. We also empirically observed that the validation \lov{}-Softmax loss closely correlated with the validation mean IoU: the epoch for which the validation loss reached its minimum was consistently close to that with the maximum mIoU, a desirable property not apparent for cross-entropy-based losses.

\noindent\tbf{Oversampling:}
To demonstrate the effect of oversampling in the presence of extreme class imbalance we separately report the mean IoU over anatomical, tool and rare classes, the latter being tool classes occurring in less than $10\%$ of records. As shown in Tab.~\ref{tab:rf}, training with either RF or Adaptive sampling significantly boosts performance on rare and tool classes, regardless of architecture.

\begin{table}[t]
\setlength{\tabcolsep}{4pt}
  \centering
  \caption{Effect of oversampling in \ta2 and  \ta3 (all use \lov{} loss)}
  \label{tab:rf}
  \begin{tabular}{*9c}
    \toprule
    \multicolumn{3}{c}{Model}   & \multicolumn{2}{c}{Oversampling} & \multicolumn{4}{c}{Mean IoU}  \\
    \cmidrule(r){1-3} \cmidrule(l){4-5} \cmidrule(l){6-8} \cmidrule(l){9-9}
    Task       & Backbone         & Decoder       & Adaptive    & RF         & Anatomies   & Tools       & Rare        & Overall \\
    \mc{\ta2} & \mc{ResNet$50$}  & \mc{OCR}      & \mc{}       & \mc{}      & \mc{0.9038} & \mc{0.7502} & \mc{0.7572} & \mc{0.8109} \\
    \mc{\ta2} & \mc{ResNet$50$}  & \mc{OCR}      & \mc{\tick}  & \mc{}      & \mc{0.9117} & \mc{0.7666} & \mc{0.7614} & \mc{0.8220}\\
    \mc{\ta2} & \mc{ResNet$50$}  & \mc{OCR}      & \mc{}       & \mc{\tick} & \mc{0.9063} & \mc{0.7689} & \mc{0.7752} & \mc{0.8236}\\

    \midrule
    \mc{\ta2} & \mc{ResNet$34$}  & \mc{\upr}     & \mc{}       & \mc{}      & \mc{0.9061} & \mc{0.7546} & \mc{0.7374} & \mc{0.8151} \\
    \mc{\ta2} & \mc{ResNet$34$}  & \mc{\upr}     & \mc{\tick}  & \mc{}      & \mc{0.9045} & \mc{0.7620} & \mc{0.7665} & \mc{0.8198} \\
    \mc{\ta2} & \mc{ResNet$34$}  & \mc{\upr}     & \mc{}       & \mc{\tick} & \mc{0.9045} & \mc{0.7787} & \mc{0.7836} & \mc{0.8298} \\
    \midrule
    \mc{\ta3} & \mc{ResNet$50$}  & \mc{OCR}      & \mc{}       & \mc{}      & \mc{0.9062} & \mc{0.6695} & \mc{0.6202} & \mc{0.7345}\\
    \mc{\ta3} & \mc{ResNet$50$}  & \mc{OCR}      & \mc{\tick}  & \mc{}      & \mc{0.9020} & \mc{0.7112} & \mc{0.6913} & \mc{0.7622} \\
    \mc{\ta3} & \mc{ResNet$50$}  & \mc{OCR}      & \mc{}  & \mc{\tick}      & \mc{0.9059} & \mc{0.7296} & \mc{0.7144} & \mc{0.7777}  \\

    \midrule
    \mc{\ta3} & \mc{ResNet$34$}  & \mc{\upr}     & \mc{}       & \mc{}      & \mc{0.9031} & \mc{0.6744} & \mc{0.6274} & \mc{0.7374} \\
    \mc{\ta3} & \mc{ResNet$34$}  & \mc{\upr}     & \mc{\tick}  & \mc{}      & \mc{0.9010} & \mc{0.6860} & \mc{0.6571} & \mc{0.7457} \\
    \mc{\ta3} & \mc{ResNet$34$}  & \mc{\upr}     & \mc{}       & \mc{\tick} & \mc{0.9024} & \mc{0.7238} & \mc{0.7076} & \mc{0.7735} \\
    \bottomrule
  \end{tabular}
\end{table}

\noindent\tbf{Backbone effect:}
Against a baseline of $0.8109$ mIoU at \ta2 (\upr{} head, ResNet$34$ backbone, \lov{}-Softmax loss, no augmentations or oversampling), we found most deeper backbones such as ResNet$50$, WideResNet$50$, WideResNet$101$ to yield no returns. Consistent with \cite{resnext}, only ResNeXt$50$ and ResNeXt$101$ were promising with respective mIoU values of $0.8132$ and $0.8195$, at the cost of higher GPU memory usage. Furthermore, initialising OCRNet's ResNet$50$ backbone with MoCov2 \cite{MoCov2} instead of Imagenet weights slightly boosted performance as shown in Tab.~\ref{tab:results}, hinting that transfer learning from natural to surgical scenes from a self-supervised initialisation can be more effective.

\noindent\tbf{Comparison with existing work:}
Our results outperform the state-of-the-art on the CaDIS benchmark established in \cite{CaDIS}. For our previous experiments we utilised a filtered version of the dataset as described in Sec.~\ref{methods:data}. For a direct comparison, we report results using the train-validation-test split of \cite{CaDIS}, without filtering out mislabelled frames. As shown in Tab.~\ref{tab:comparison} our methodology outperforms results of \cite{CaDIS} across all tasks and network architectures by a large margin. Notably, the gain in performance by our methodology increases for \ta2 and \ta3 that present high imbalance in the class distributions, which can be attributed to the fact that we employ oversampling methods that explicitly address it.

\begin{table}[t]
\setlength{\tabcolsep}{2.5pt}
  \centering
  \caption{Results on train-val-test dataset split of \cite{CaDIS}, \tbf{bold} denotes the best result per model, \br{red} and \bb{blue} the best overall in val and test set of each task respectively.}
  \label{tab:comparison}
  \begin{tabular}{ccccccccc}
    \toprule
    \mc{Model} & \multicolumn{2}{c}{Training} & \multicolumn{2}{c}{\ta1} & \multicolumn{2}{c}{\ta2} & \multicolumn{2}{c}{\ta3} \\
    \cmidrule(l){1-1} \cmidrule(l){2-3} \cmidrule(l){4-5} \cmidrule(l){6-7} \cmidrule(l){8-9}
    \mc{}        & Loss        & Sampling & val           & test          & val            & test           & val            & test\\
    \mc{\dvp{} \cite{CaDIS}} & \mc{CE}     & \mc{-}   & \mc{$0.8530$} & \mc{$0.8262$} & \mc{$0.7450$}  & \mc{$0.7226$}  & \mc{$0.6860$}  & \mc{$0.6323$}\\
    \mc{\dvp{}} & \mc{\lov{}} & \mc{RF}  & \mc{\tbf{0.8848}}  & \mc{\tbf{0.8565}}  & \mc{\tbf{0.7914}}  & \mc{\tbf{0.7517}}  & \mc{\tbf{0.7744}}  & \mc{\tbf{0.7051}}\\
   \midrule
    \mc{\upr{} \cite{CaDIS}}  & \mc{CE}     & \mc{-}   & \mc{$0.8790$} & \mc{$0.8396$}  & \mc{$0.7950$}  & \mc{$0.7376$}  & \mc{$0.7420$}  & \mc{$0.6676$}\\
    \mc{\upr{}} & \mc{\lov{}} & \mc{RF}  & \mc{\tbf{0.8885}}  & \mc{\tbf{0.8632}}  & \mc{\tbf{0.8154}}  & \mc{\tbf{0.7688}}  & \mc{\tbf{0.7575}}  & \mc{\tbf{0.7044}}\\
    \midrule
    \mc{HRNetv2 \cite{CaDIS}} & \mc{CE}     & \mc{-}   & \mc{$0.8810$} & \mc{$0.8491$}  & \mc{$0.8180$} & \mc{$0.7611$}  & \mc{$0.7240$}  & \mc{$0.6664$}\\
    \midrule
    \mc{OCR}    & \mc{\lov{}} & \mc{RF}  & \mc{\br{0.8897}}  & \mc{\bb{0.8640}}   & \mc{\br{0.8325}} & \mc{\bb{0.7909}}  & \mc{\br{0.7940}}  & \bb{0.7194}\\
    \bottomrule
  \end{tabular}
\end{table}

\section{Conclusion}\label{discussion}
The CaDIS dataset is a new benchmark in the domain of cataract surgery. Its main challenge lies in the extreme class imbalance encountered in the highly granular semantic segmentation labels provided, which we meet with two different data oversampling strategies. We set a new state-of-the-art and provide extensive ablation studies on the effect of different design choices. These show that the effect of varying encoder or decoder designs is minor and generally difficult to predict, while the choice of the loss function and data sampling strategy are paramount. Specifically, we recommend the use of the \lov{}-Softmax loss as a differentiable surrogate for the Jaccard index \cite{Lovasz}, and an adaptation of repeat factor sampling \cite{gupta2019lvis} to increase the frequency of hard records in the training data. Our findings can be applied to other datasets with a similar class imbalance, and may guide efforts of pushing the state-of-the-art further on CaDIS.

\subsubsection{Acknowledgements}
The authors would like to thank Martin Huber, Jeremy Birch and Joan M.~Nunez Do Rio for their contributions in the EndoVIS challenge participation. This work was supported by the National Institute for Health Research NIHR (Invention for Innovation, i4i; II-LB-0716-20002). The views expressed are those of the authors and not necessarily those of the NHS, the NIHR, or the Department of Health and Social Care.

\bibliographystyle{splncs04.bst}
\bibliography{ref}

\begin{thebibliography}{10}
\providecommand{\url}[1]{\texttt{#1}}
\providecommand{\urlprefix}{URL }
\providecommand{\doi}[1]{https://doi.org/#1}

\bibitem{WHO_blindness}
Blindness and vision impairment.
  \url{https://www.who.int/news-room/fact-sheets/detail/blindness-and-visual-impairment},
  accessed: 2021-03-01

\bibitem{CATARACTS_workflow}
Al~Hajj, H., Lamard, M., Conze, P.H., Roychowdhury, S., Hu, X.,
  Mar{\v{s}}alkait{\.e}, G., Zisimopoulos, O., Dedmari, M.A., Zhao, F.,
  Prellberg, J., et~al.: Cataracts: Challenge on automatic tool annotation for
  cataract surgery. Medical image analysis  \textbf{52},  24--41 (2019)

\bibitem{berger2018adaptive}
Berger, L., Eoin, H., Cardoso, M.J., Ourselin, S.: An adaptive sampling scheme
  to efficiently train fully convolutional networks for semantic segmentation.
  In: Annual Conference on Medical Image Understanding and Analysis. pp.
  277--286. Springer (2018)

\bibitem{Lovasz}
Berman, M., Triki, A.R., Blaschko, M.B.: The lov{\'a}sz-softmax loss: A
  tractable surrogate for the optimization of the intersection-over-union
  measure in neural networks. In: Proceedings of the IEEE Conference on
  Computer Vision and Pattern Recognition. pp. 4413--4421 (2018)

\bibitem{Deeplabv3plus}
Chen, L.C., Zhu, Y., Papandreou, G., Schroff, F., Adam, H.: Encoder-decoder
  with atrous separable convolution for semantic image segmentation. In:
  Proceedings of the European conference on computer vision (ECCV). pp.
  801--818 (2018)

\bibitem{MoCov2}
Chen, X., Fan, H., Girshick, R., He, K.: Improved baselines with momentum
  contrastive learning. arXiv preprint arXiv:2003.04297  (2020)

\bibitem{CaDIS_tool_segmentation}
Fox, M., Taschwer, M., Schoeffmann, K.: Pixel-based tool segmentation in
  cataract surgery videos with mask r-cnn. In: 2020 IEEE 33rd International
  Symposium on Computer-Based Medical Systems (CBMS). pp. 565--568. IEEE (2020)

\bibitem{CaDIS}
Grammatikopoulou, M., Flouty, E., Kadkhodamohammadi, A., Quellec, G., Chow, A.,
  Nehme, J., Luengo, I., Stoyanov, D.: Cadis: Cataract dataset for surgical
  rgb-image segmentation. Medical Image Analysis  \textbf{71},  102053 (2021).
  \doi{https://doi.org/10.1016/j.media.2021.102053}

\bibitem{gupta2019lvis}
Gupta, A., Dollar, P., Girshick, R.: Lvis: A dataset for large vocabulary
  instance segmentation. In: Proceedings of the IEEE Conference on Computer
  Vision and Pattern Recognition. pp. 5356--5364 (2019)

\bibitem{ResNet}
He, K., Zhang, X., Ren, S., Sun, J.: Deep residual learning for image
  recognition. In: Proceedings of the IEEE conference on computer vision and
  pattern recognition. pp. 770--778 (2016)

\bibitem{karpathy_recipe}
Karpathy, A.: A recipe for training neural networks (2019),
  http://karpathy.github.io/2019/04/25/recipe/

\bibitem{Adam}
Kingma, D.P., Ba, J.: Adam: {A} method for stochastic optimization. In: Bengio,
  Y., LeCun, Y. (eds.) 3rd International Conference on Learning
  Representations, {ICLR} 2015 (2015)

\bibitem{SDSc2020}
Maier-Hein, L., Eisenmann, M., Sarikaya, D., M{\"a}rz, K., Collins, T.,
  Malpani, A., Fallert, J., Feussner, H., Giannarou, S., Mascagni, P., et~al.:
  Surgical data science--from concepts to clinical translation. arXiv preprint
  arXiv:2011.02284  (2020)

\bibitem{torchvision}
Marcel, S., Rodriguez, Y.: Torchvision the machine-vision package of torch. In:
  Proceedings of the 18th ACM international conference on Multimedia. pp.
  1485--1488 (2010)

\bibitem{polynomial}
Mishra, P., Sarawadekar, K.: Polynomial learning rate policy with warm restart
  for deep neural network. In: TENCON 2019-2019 IEEE Region 10 Conference
  (TENCON). pp. 2087--2092. IEEE (2019)

\bibitem{MoritaPhase}
Morita, S., Tabuchi, H., Masumoto, H., Yamauchi, T., Kamiura, N.: Real-time
  extraction of important surgical phases in cataract surgery videos.
  Scientific reports  \textbf{9}(1), ~1--8 (2019)

\bibitem{CataractTools1}
Ni, Z.L., Bian, G.B., Zhou, X.H., Hou, Z.G., Xie, X.L., Wang, C., Zhou, Y.J.,
  Li, R.Q., Li, Z.: Raunet: Residual attention u-net for semantic segmentation
  of cataract surgical instruments. In: International Conference on Neural
  Information Processing. pp. 139--149. Springer (2019)

\bibitem{Padoy2012}
Padoy, N., Blum, T., Ahmadi, S.A., Feussner, H., Berger, M.O., Navab, N.:
  Statistical modeling and recognition of surgical workflow. Medical image
  analysis  \textbf{16}(3),  632--641 (2012)

\bibitem{ohem}
Shrivastava, A., Gupta, A., Girshick, R.: Training region-based object
  detectors with online hard example mining. In: Proceedings of the IEEE
  Conference on Computer Vision and Pattern Recognition (CVPR) (June 2016)

\bibitem{ohem_code}
Sun, K., Xiao, B., Liu, D., Wang, J.: Deep high-resolution representation
  learning for human pose estimation. In: CVPR (2019)

\bibitem{wang2016cataract}
Wang, W., Yan, W., Fotis, K., Prasad, N.M., Lansingh, V.C., Taylor, H.R.,
  Finger, R.P., Facciolo, D., He, M.: Cataract surgical rate and
  socioeconomics: a global study. Investigative ophthalmology \& visual science
   \textbf{57}(14),  5872--5881 (2016)

\bibitem{UPerNet}
Xiao, T., Liu, Y., Zhou, B., Jiang, Y., Sun, J.: Unified perceptual parsing for
  scene understanding. In: Proceedings of the European Conference on Computer
  Vision (ECCV). pp. 418--434 (2018)

\bibitem{resnext}
Xie, S., Girshick, R., Doll{\'a}r, P., Tu, Z., He, K.: Aggregated residual
  transformations for deep neural networks. In: Proceedings of the IEEE
  conference on computer vision and pattern recognition. pp. 1492--1500 (2017)

\bibitem{PhaseJAMA}
Yu, F., Croso, G.S., Kim, T.S., Song, Z., Parker, F., Hager, G.D., Reiter, A.,
  Vedula, S.S., Ali, H., Sikder, S.: Assessment of automated identification of
  phases in videos of cataract surgery using machine learning and deep learning
  techniques. JAMA network open  \textbf{2}(4),  e191860--e191860 (2019)

\bibitem{OCR}
Yuan, Y., Chen, X., Wang, J.: Object-contextual representations for semantic
  segmentation. arXiv preprint arXiv:1909.11065  (2019)

\bibitem{CataractTools2}
Zang, D., Bian, G.B., Wang, Y., Li, Z.: An extremely fast and precise
  convolutional neural network for recognition and localization of cataract
  surgical tools. In: International Conference on Medical Image Computing and
  Computer-Assisted Intervention. pp. 56--64. Springer (2019)

\bibitem{zhou2017scene}
Zhou, B., Zhao, H., Puig, X., Fidler, S., Barriuso, A., Torralba, A.: Scene
  parsing through ade20k dataset. In: Proceedings of the IEEE Conference on
  Computer Vision and Pattern Recognition (2017)

\bibitem{zhou2018semantic}
Zhou, B., Zhao, H., Puig, X., Xiao, T., Fidler, S., Barriuso, A., Torralba, A.:
  Semantic understanding of scenes through the ade20k dataset. International
  Journal on Computer Vision  (2018)

\bibitem{DeepPhase}
Zisimopoulos, O., Flouty, E., Luengo, I., Giataganas, P., Nehme, J., Chow, A.,
  Stoyanov, D.: Deepphase: surgical phase recognition in cataracts videos. In:
  International Conference on Medical Image Computing and Computer-Assisted
  Intervention. pp. 265--272. Springer (2018)

\end{thebibliography}

\end{document}